%% file: main.tex
\PassOptionsToPackage{dvipdfmx}{graphicx}
\documentclass[sigconf]{acmart}
 
\def\BibTeX{{\rm B\kern-.05em{\sc i\kern-.025em b}\kern-.08emT\kern-.1667em\lower.7ex\hbox{E}\kern-.125emX}}

\usepackage{xcolor}

\usepackage{nicefrac}
\usepackage{siunitx}
\usepackage{array,framed}
\usepackage{booktabs}
\usepackage{multirow}
\usepackage{
  color,
  float,
  epsfig,
  wrapfig,
  graphics,
  graphicx,
  subcaption
}
\usepackage{textcomp,amssymb}
\usepackage{setspace}
\usepackage{latexsym,fancyhdr,url}
\usepackage{enumerate}
\usepackage{algpseudocode}
\usepackage{graphics}
\usepackage{xparse} 
\usepackage{xspace}
\usepackage{multirow}
\usepackage{csvsimple}
\usepackage{balance}
\usepackage{algorithm}
\usepackage{subcaption}

\usepackage{
  tikz,
  pgfplots,
  pgfplotstable
}
\usepackage{hyperref}

\usetikzlibrary{
  shapes.geometric,
  arrows,
  external,
  pgfplots.groupplots,
  matrix
}

\pgfplotsset{compat=1.9}

\usepackage{mathtools,}

\DeclareMathAlphabet{\mathcal}{OMS}{cmsy}{m}{n}

\DeclareGraphicsExtensions{%
    .png,.PNG,%
    .pdf,.PDF,%
    .jpg,.mps,.jpeg,.jbig2,.jb2,.JPG,.JPEG,.JBIG2,.JB2}

\input{defs}
\begin{document}
\fancyhead{}
\def\thetitle{Adversarial Attacks on Hidden Tasks in Multi-Task Learning}

\title{\thetitle}

\author{Yu Zhe}
\affiliation{%
  \institution{RIKEN AIP}
  \city{Tokyo}
}

\author{Rei Nagaike}
\affiliation{%
  \institution{Tsukuba university}
  \city{Tsukuba}
}

\author{Daiki Nishiyama}
\affiliation{%
  \institution{RIKEN AIP/ Tokyo Institute of Technology}
  \city{Tokyo}
}

\author{Kazuto Fukuchi}
\affiliation{%
  \institution{RIKEN AIP/Tsukuba university}
  \city{Tokyo/Tsukuba}
}

\author{Jun Sakuma}
\affiliation{%
  \institution{RIKEN AIP/ Tokyo Institute of Technology}
  \city{Tokyo}
}

\date{}
\begin{abstract}
Deep learning models are susceptible to adversarial attacks, where slight perturbations to input data lead to misclassification. Adversarial attacks become increasingly effective with access to information about the targeted classifier. In the context of multi-task learning, where a single model learns multiple tasks simultaneously, attackers may aim to exploit vulnerabilities in specific tasks with limited information.  This paper investigates the feasibility of attacking hidden tasks within multi-task classifiers, where model access regarding the hidden target task and labeled data for the hidden target task are not available, but model access regarding the non-target tasks is available. We propose a novel adversarial attack method that leverages knowledge from non-target tasks and the shared backbone network of the multi-task model to force the model to forget knowledge related to the target task. Experimental results on CelebA and DeepFashion datasets demonstrate the effectiveness of our method in degrading the accuracy of hidden tasks while preserving the performance of visible tasks, contributing to the understanding of adversarial vulnerabilities in multi-task classifiers.
\end{abstract}

\maketitle

\section{Introduction}
\label{sec:intro}

Deep learning is known to be vulnerable to adversarial attacks \cite{goodfellow2014explaining}. Adversarial attacks involve perturbing input samples, known as adversarial samples, fed into a trained classifier to induce misclassification \cite{goodfellow2014explaining}. The effectiveness of adversarial attacks usually increases as the attacker gains knowledge about the output, architecture, parameters, and other aspects of the targeted classifier.

\begin{figure}[!t]
    \centering
    \includegraphics[width=\columnwidth]{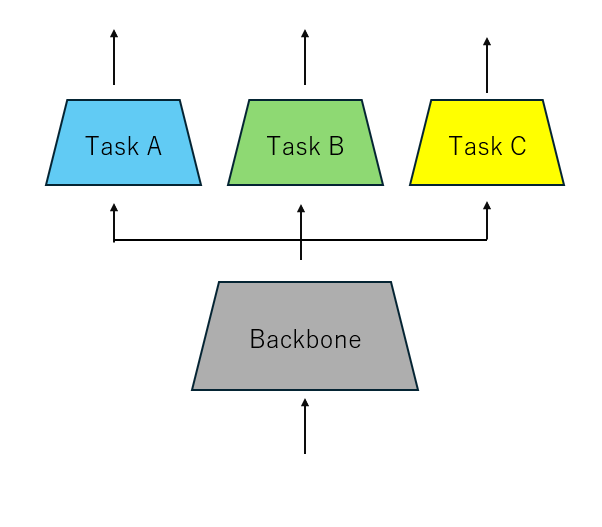}
    \caption{The structure of a typical multi-task classifier. The gray part represents the backbone network $B(\circ)$, which involves model parameters shared across all tasks, while the blue, green, and yellow parts represent the headers for each task.}
    \label{fig:multi-task}
\end{figure}

In contrast to single-task learning, where one classifier learns one task, multi-task learning, which involves learning multiple tasks simultaneously, has been reported to exhibit superior performance by leveraging knowledge from multiple tasks \cite{vandenhende2021multi}. For instance, multi-task learning has been successfully applied in fields such as autonomous driving, which involves object detection and semantic segmentation on forward-facing car images \cite{chen2018multi}, multi-task learning has been successfully applied. In this scenario, a multi-task model can simultaneously learn to detect objects on the road while also segmenting the scene for navigation purposes, leading to more robust and accurate performance compared to single-task approaches. Additionally, in facial recognition \cite{ranjan2017hyperface}, which includes tasks like gender and individual estimation, multi-task learning has been reported to outperform single-task learning. By jointly learning to recognize gender and individual identities from facial images, a multi-task model can benefit from shared representations, improving the accuracy of both tasks. The general architecture of multi-task models consists of a shared feature extractor (backbone network) and task-specific headers connected to it, as Figure \ref{fig:multi-task} shows.

When constructing a multi-task model, pre-trained models are often used as the backbone. In recent years, a large number of pre-trained models with excellent performance are available as open-source models. For example, in the field of imagery, ResNet \cite{he2016deep}, VGG \cite{simonyan2015very}, Inception (GoogLeNet) \cite{szegedy2015going}, MobileNet \cite{howard2017mobilenets}, EfficientNet \cite{tan2019efficientnet}, etc. are known. In the field of natural language, BERT \cite{devlin2018bert}, GPT-2 \cite{radford2019language}, GPT-3 \cite{brown2020language}, XLNet \cite{yang2019xlnet}, RoBERTa \cite{liu2019roberta}, T5 \cite{raffel2019exploring} and others are available as open-source models. All of these models are widely used through GitHub and Hugging Face, where the model parameters themselves are publicly available. 
Model building with such pre-learning models as a backbone is widely used both in research and in industry.

We consider a scenario where an attacker targets a specific task within the multitask framework. The attacker aims to attack one specific task (the target task) among the multiple tasks. In this scenario, we further assume that while information about the non-target tasks can be obtained through white-box or black-box access, neither information about the target task nor training data for the target task is accessible. The following raises scenarios of multitask classification where the presence of specific tasks is hidden.

\vspace{0.25cm}

{\bf Scenario 1. Individual identification.}
Suppose a facial authentication system at the entrance of a facility, which identifies individuals and grants entry. In addition, suppose that this system secretly estimates the gender and age of individuals. In this case, anyone would be apparently aware that the system implements the functionality of individual identification while the functionality of personal attribute estimation, such as gender and age determination, for collecting visitor attributes may not be disclosed to users. 

\vspace{0.25cm}

{\bf Scenario 2. Smartphone voice commands.}
Consider a voice command detector in a smartphone. This detector constantly monitors the surrounding audio and reacts to certain trigger words by users. In addition, suppose that this detector can secretly estimate speaker attributes (e.g., gender and age) and some information collected from the environment (e.g., the number of family members and conversation topics). In this case, everyone would know that this detector has a voice command detection function, but the other functionalities would not be revealed to the user.

\vspace{0.25cm}

The aim of this study is to investigate the following research questions: 
\begin{itemize}
\item RQ1 (attack performance). For such a multi-task classifier involving tasks that are inaccessible from the attacker's side, is it possible to attack the invisible tasks and invalidate their functionality? 
\item RQ2 (stealthiness). Is it possible to carry out the attack without affecting the performance of the tasks that are visible from the attacker's side?
\end{itemize}

Adversarial attack methods are often designed assuming that some information (e.g., models in the white-box case and model outputs in the black-box case) are obtainable.
Therefore, when information about the target task is unavailable, most of existing adversarial attack methods are unavailable.
Existing methods that can be applied to such situations are the attack method designed for the no-box setting, which attacks the classifier without relying on assumptions or knowledge about the target task \cite{chen2017zoo}.
Although the attack method in the no-box setting can be used to attack invisible tasks in a multitask classifier, the impact of attacks significantly decreases compared to situations where tasks are visible (RQ1).

Stealthiness of attacks is also crucial in this study (RQ2). In multi-task learning, improving stealthiness can be achieved by ensuring that tasks other than the target task are not adversely affected. For example, in the aforementioned facial recognition system, while accurate decisions regarding entry permissions are desired, tasks such as gender prediction may need to be misled to maintain privacy. Existing attacks, such as no-box attacks, are not stealthy because they are not tailored to selectively affect invisible tasks, causing performance degradation of visible tasks (RQ2).

Assuming the classification model follows the multi-task model architecture in Fig. \ref{fig:multi-task}, this study focuses on the threat model\footnote{Section \ref{sec:threat} describes a detailed definition of our threat model.} as follows:
\begin{enumerate}
\item attackers can obtain neither information about the invisible tasks (target task) nor labeled data related to the target task;
\item however, the attackers can have white-box access to the backbone network and white-box or black-box access to the header network for visible tasks (non-target tasks). 
\end{enumerate}


As already introduced, multi-task model building with publicly available pre-trained models as a backbone is widely used.
For example, pre-trained ResNet-18 \cite{he2016deep} on ImageNet \cite{deng2009imagenet} is cited more than 10,0000 times and used as the basis for image classification.
Also, the pre-trained language model BERT\cite{devlin2018bert} is cited nearly 100,000 times and is often used as the basis for natural language processing, too.
Fine-tuning of pre-trained models is relatively laborious and expensive, so many researchers and developers do not fine-tune these pre-trained models, but use them as fixed feature extractors and train task-specific heads only in many cases.
For these reasons, the threat model discussed above can be readily realized if model developers use pre-trained models without any modification and let users have white-box or black-box access to the heads of the multi-task model.



In this study, our aim is to conduct effective attacks on hidden target tasks (RQ1), and we propose adversarial attack methods that utilize information from tasks other than the hidden task when the backbone network is available to the attacker. Due to the structure of multi-task classifiers, which share knowledge between tasks, even if information about the target task is not accessible, acquiring information about other tasks or the backbone network can lead to the development of effective attacks on the target task. Specifically, we propose attacks focusing on the output (feature vectors) of the backbone network, which shares knowledge between tasks. On the other hand, we consider attacks on hidden tasks while preserving the accuracy of other tasks (RQ2).


\subsection{Related Work}\label{sec:related}
\textbf{Adversarial attacks on the non-hidden tasks against single-task classifiers.} Adversarial attack methods, such as FGSM \cite{goodfellow2014explaining}, PGD attack \cite{madry2017towards}, and CW attack \cite{carlini2017towards}, have been proposed when information such as gradients of the loss with respect to the input for the target task is available. Additionally, some attack methods have been proposed when only the output of the target classifier is available. Under this setting, some attacks transfer adversarial samples generated by surrogate classifiers mimicking the target classifier to the target classifier \cite{papernot2017practical, lord2022attacking}; some attacks optimize adversarial perturbation by observing the change of output from the target model without surrogate models \cite{guo2019simple}. However, these attacks become infeasible when it is impossible to obtain information about the target task for training surrogate models or computing gradients.

\textbf{Adversarial attacks on the hidden tasks of single-task classifiers.} Adversarial attacks that are feasible even when no information about the target classifier, including gradient, output, etc, is available are referred to as no-box attacks \cite{chen2017zoo}. Research on no-box attacks against single-task classifiers has been extensively conducted \cite{li2020practical, lu2023hard, bose2020adversarial}. These attacks first generate surrogate models by supervised or unsupervised learning using a small amount of data for a task that may be different from the target task, then generate adversarial examples for this model. The transferability of the adversarial samples allows the adversarial samples on this model to be valid on the no-box target model.

As for attack methods with excellent transferability to classifiers with different tasks or architectures, attacks focusing on the backbone network have also been studied \cite{lu2020enhancing, naseer2018task, wang2021feature}. These attacks alter output features of the backbone network in various ways, disrupting the outputs of specific layers and subsequent layers, thereby degrading accuracy. These attack techniques are effective regardless of the task and, therefore, have an attack effect on hidden tasks. On the other hand, for the same reason, they also have an attack effect on nonhidden tasks and are therefore not stealthy.

\textbf{Adversarial attacks on multi-task classifiers} have been studied extensively. Mao et al. \cite{mao2020multitask, ghamizi2022adversarial} applied adversarial attacks designed for single-task classifiers to multitask classifiers and compared the vulnerabilities between single-task and multi-task classifiers. Sobh et al. \cite{gurulingan2021uninet, sobh2021adversarial} demonstrated that attacking one task of a multitask classifier leads to a decrease in predictive performance for other tasks as well. Guo et al. \cite{guo2020multi} proposed attacks deceiving multiple tasks simultaneously using a multitask generative model to generate adversarial perturbations for multiple tasks. In existing work, it is difficult to reduce the impact on tasks other than the target task. As discussed above, some attacks are not feasible due to their threat model differing from the attacks considered in this study. Methods such as no-box attacks, while they can attack hidden tasks, they can not only affect the target task only.

\begin{figure*}[!t]
    \centering
    \includegraphics[width=\textwidth]{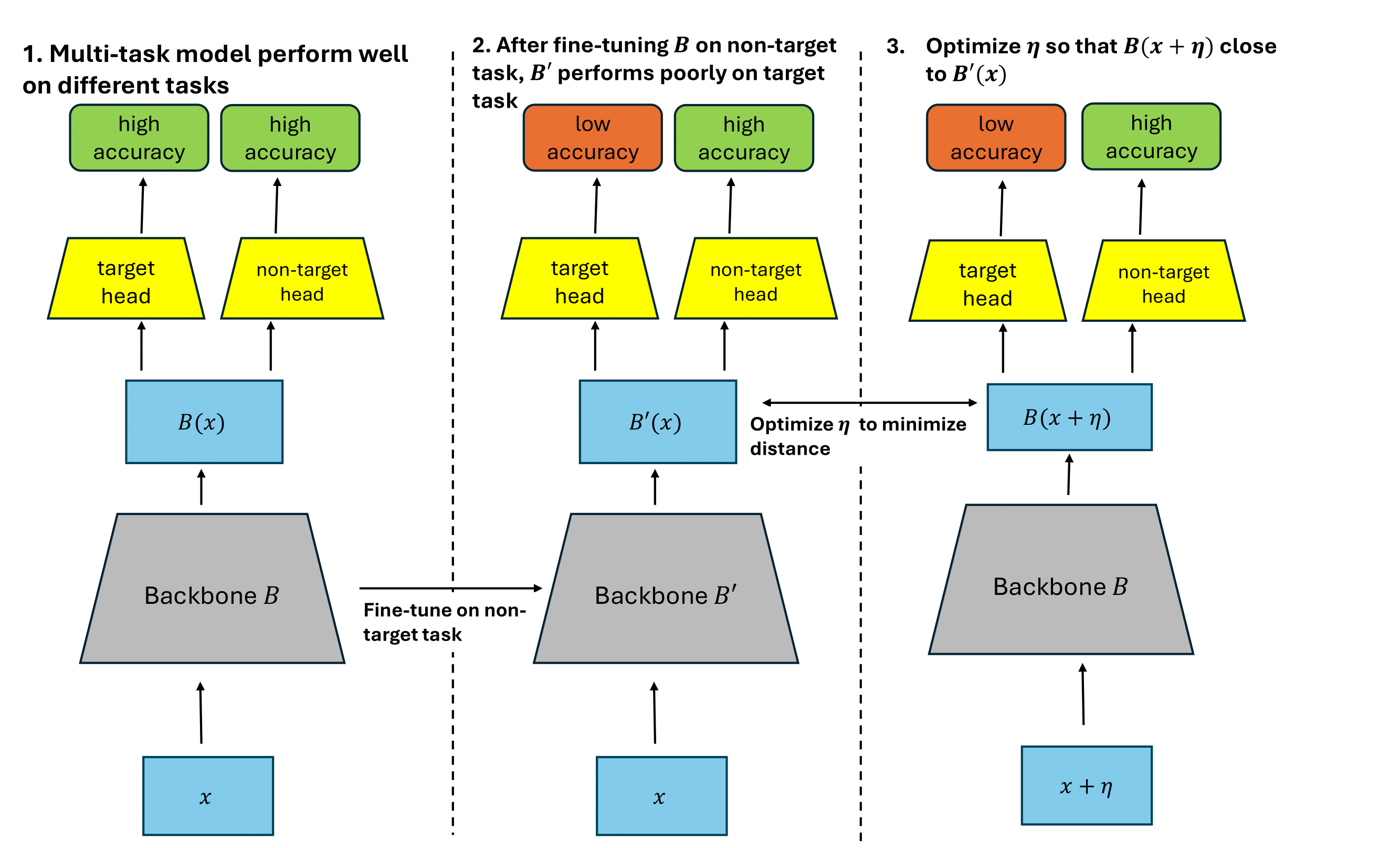}
    \caption{The illustration of proposed CF attack. We first fine-tune $B$ on non-target tasks only and get $B'$. $B'$ is not able to extract features related to the target task. We aim to design adversarial perturbation $\eta$ so that $B(x+\eta)$ is close to $B'(x)$. Such features will cause the target head to give the wrong prediction.}
    \label{fig: cf_attack}
\end{figure*}

\subsection{Contribution}\label{sec:contri}
To frame our contributions effectively, we first revisit the specific characteristics of the problem at hand: (1) crafting adversarial examples that work effectively for the target task inaccessible to attackers; (2) crafting adversarial examples solely inducing performance degradation in the target task, while leaving other tasks unaffected. To achieve this, we leverage catastrophic forgetting \cite{french1999catastrophic}, a phenomenon in model training where knowledge of one task is lost after training with others. 

The idea of our attack consists of the following 3-steps. 
\begin{enumerate}
\item The attacker finetunes the well-trained multi-task classifier (Fig. \ref{fig: cf_attack} left) with data not containing data about the hidden target task to induce catastrophic forgetting. (Fig. \ref{fig: cf_attack} center). As a result, the multi-task classifier loses the ability to deal with the hidden target task. 
\item Next, the attacker obtains the feature vector of the original backbone $B(x)$ and the backbone after catastrophic forgetting $B'(x)$. Here, $B(x)$ can deal with the target task while $B'(x)$ cannot due to the catastrophic forgetting.
\item The attacker finds perturbation $\eta$ so that $B'(x)$ is reproduced with the original backbone $B$. Specifically, $\eta$ such that $\|B(x+\eta) - B'(x)\|$ takes small values is obtained by optimization (Fig. \ref{fig: cf_attack} right).
\end{enumerate}

Note that $B'(x)$ is affected by catastrophic forgetting, which only degrades the prediction performance of the hidden target task. Thus, $B(x+\eta)$ has low predictability in the hidden target task (RQ1) but little performance degradation in the other visible non-target tasks due to finetuning (RQ2).


To sum up, the contributions of this study are as follows:
\begin{enumerate}
\item This study proposes CF attack for attacking hidden tasks in multitask classifiers, where knowledge of the classifier processing the target task is not available, unlike existing adversarial attack methods. However, information about classifiers processing tasks other than the target task is accessible in the threat model considered. 

\item While existing adversarial attack methods is able to attack all tasks simultaneously, they often do not consider how to suppress the impact on tasks other than the target task. Suppressing the impact of attacks on tasks other than the target task can improve the stealthiness of attacks and make them harder to detect. We mitigate the impact on non-hidden tasks by adding a penalty so that the adversarial features are still good features for non-target tasks.

\item We conduct experiments on two multi-task datasets, CelebA \cite{celeb} and DeepFashion \cite{deepfashion}. Through experiments, we observe that our attack leads to at least 20\% performance degradation on target tasks compared to existing no-box attacks, and, at the same time, our proposal only has a limited impact on non-target tasks. 

\end{enumerate}

\section{Preliminary}
\label{sec:preliminary}

\subsection{Supervised Learning}
A single-task classifier with parameters $\theta$ is denoted as $F^{\theta}: \mathbb{R}^d \rightarrow\{1, \ldots, C\}$, where the dataset is $D=\{(\boldsymbol{x}, y)\}$, with $x$ is a random variable on $\mathbb{R}^d$ and $y$ is a random variable on $\{1, \ldots, C\}$. In this context, learning methods that aim to maximize prediction accuracy using samples and their corresponding ground truths $(\boldsymbol{x},y)$ drawn from $D$ are referred to as supervised learning:
$$\max _\theta \sum_{(\boldsymbol{x}, y) \sim D} \mathbf{1}\left[F^\theta(\boldsymbol{x})=y\right]$$
To maximize prediction accuracy, the parameters $\theta$ of the model $F^{\theta}$ are updated to minimize the loss function $L$: $$\min _\theta \sum_{(\boldsymbol{x}, y) \sim D} \mathcal{L}\left(F^\theta(\boldsymbol{x}), y\right)$$ For brevity, the notation for the parameters $\theta$ of each classifier will be omitted hereafter.

\subsection{Multi-task learning}\label{sec:pre-multi}
Multi-task learning deals with multiple takes. Let the set of tasks be $T=\left\{t_1, \ldots, t_{|T|}\right\}$. $t_i \in T$ is a task that performs $C_{i}$-class classification. Let $D_T=\{(\boldsymbol{x}, \boldsymbol{y})\}$ be a dataset that includes all tasks in $T$. Here, $\boldsymbol{y}$ is a vector containing gound-truth elements for $t_{i} \in T$ for the input sample $\boldsymbol{x}$, denoted as $\boldsymbol{y}=\left\{y_1, \ldots,\right\}$. Throughout this study, we consistently treat all task groups as classification problems. We denote the classifier trained to solve the task group $T$ (multi-task classifier) as $F_{T}$, and the classifier corresponding to the classification of $t_{i}$ as $F_i: \mathbb{R}^d \rightarrow\left\{1, \ldots, C_i\right\}$. The objective of multi-task learning can be to maximize the predictive accuracy of all tasks or to maximize the predictive accuracy of one or some tasks. In this study, we consider the former case, where the predictive accuracy of all tasks is maximized:
$$\max \sum_{i=1}^{|T|} \sum_{\left(\boldsymbol{x},\left\{y_1, \cdots, y_{|T|}\right\}\right) \sim D_T} \mathbf{1}\left[F_i(\boldsymbol{x})=y_i\right]$$

Figure \ref{fig:multi-task} illustrates a common architecture used in multi-task learning \cite{chen2018multi}. It consists of a backbone network $B(\cdot)$, task-specific header $H_i(\cdot)$ for each task $t_i$ (where $i=1, \ldots,|T|$), and a set of headers $\boldsymbol{H}=\left\{H_1, \cdots, H_{|T|}\right\}$. In this architecture, the output for task $t_i$ can be expressed as $F_i(x)=H_i \circ B(\boldsymbol{x})$. The architecture shown in Figure \ref{fig:multi-task} is commonly used in prior research on adversarial attacks against multi-task classifiers \cite{mao2020multitask, sobh2021adversarial} and this study also considers this architecture for multi-task classifiers.

We then explain the objective function of multi-task learning. Let $\left.\mathcal{L}_i\left(F_i(\boldsymbol{x})\right), y_i\right)$ denotes the loss for task $t_i \in T$ with respect to a sample $\left(\boldsymbol{x}, y_1, \cdots, y_i, \cdots, y_{|T|}\right)$. In this case, the overall loss of multi-task learning is expressed as the weighted sum of the losses of all tasks, as follows:
\begin{equation}
   \mathcal{L}_T(\boldsymbol{H}, B, \boldsymbol{x}, \boldsymbol{y})=\sum_{i=1}^{|T|} \lambda_i \mathcal{L}_i\left(H_i \circ B(\boldsymbol{x}), y_i\right) 
\label{eq:multi-losses}
\end{equation}
where $\lambda_i$ is the weight for task $t_i$.

\subsection{Adversarial attack}\label{sec:pre-attack}
\textbf{Adversarial attack on single-task classifiers.} Adversarial attacks involve generating adversarial examples by adding perturbations $\delta$ to test samples $x$, leading to incorrect predictions when these samples are inputted into the machine learning models \cite{goodfellow2014explaining}. Adversarial attacks can be classified into two categories: untargeted attacks and targeted attacks. In untargeted attacks, the predicted label for the adversarial sample is not specified, and the aim is to find $\delta$ such that the predicted label for $\boldsymbol{x}+\boldsymbol{\delta}$ (i.e., $F(\boldsymbol{x}+\boldsymbol{\delta})$) differs from the true label $y$. In targeted attacks, $\boldsymbol{\delta}$ is generated so that the predicted label for the adversarial example $\boldsymbol{x}+\boldsymbol{\delta}$ becomes a specific label $c$, i.e., $F(\boldsymbol{x}+\boldsymbol{\delta})=c$ where  $\quad c \neq y$. This study focuses on untargeted attacks. The common approach for untargeted attacks on single-task classifiers involves optimizing the loss function using the sample $x$ and its corresponding ground-truth $y$ to find the adversarial perturbation $\delta$:
\begin{equation}
    \max _{\|\boldsymbol{\delta}\|_p \leq \epsilon} \mathcal{L}(F(\boldsymbol{x}+\boldsymbol{\delta}), y)
\end{equation}
where $\|\delta\|_p$ represents the $L_{p}$ norm of $\delta$. In particular, when $p=\infty$, the $\operatorname{clip}_{-\epsilon,\epsilon}$ function is used to bound the size of $\delta$ within the range $[-\epsilon, \epsilon]$. In representative adversarial attack methods like FGSM \cite{goodfellow2014explaining}, $\boldsymbol{\delta}$ is computed using the direction of the gradient with respect to the loss
\begin{equation}
    \boldsymbol{\delta}=\epsilon \operatorname{sign}\left(\nabla_{\boldsymbol{x}} \mathcal{L}(\boldsymbol{x}, y)\right)
\end{equation}
Furthermore, PGD \cite{madry2017towards} is a method that iteratively applies the perturbation generation approach of FGSM.

\begin{table*}[!t]
\caption{The information required to apply existing methods and our proposed method (ours). Model indicates whether $B$, $H_{\text{tgt}}$, and $H_{\text{no-tgt}}$ are White/Black or No-box. "sur" refers to the surrogate model acquired through available training data. Data indicates what data is needed to attack for each method.}
\scalebox{0.9}{\begin{tabular}{c|cc|c|c|c|c}
\hline
\multirow{2}{*}{Method} & \multicolumn{2}{c|}{Model}                                             & \multirow{2}{*}{Data}                                           & \multirow{2}{*}{\begin{tabular}[c]{@{}c@{}}effective when \\ $H_{\text{tgt}}$ is no-box\end{tabular}} & \multirow{2}{*}{\begin{tabular}[c]{@{}c@{}}effective when \\ $y_{\text{tgt}}$ is unavailable\end{tabular}} & \multirow{2}{*}{\begin{tabular}[c]{@{}c@{}}stealthness on\\  $t_{\text{no-tgt}}$\end{tabular}} \\ \cline{2-3}
                        & \multicolumn{1}{c|}{White/Black-box}          & No-box                 &                                                                 &                                                                                                &                                                                                                     &                                                                                         \\ \hline
FGSM \cite{goodfellow2014explaining}, PGD \cite{madry2017towards}               & \multicolumn{1}{c|}{$B, H_{\text{tgt}}$}             & $H_{\text{no-tgt}}$           &$ \left\{\left(\boldsymbol{x}, y_{\text{tgt}}\right)\right\}$            & \multirow{3}{*}{X}                                                                             & \multirow{3}{*}{X}                                                                                  & \multirow{3}{*}{X}                                                                      \\
surrogate+FGSM, PGD \cite{papernot2017practical}    & \multicolumn{1}{c|}{$B^{sur}, H_{\text{tgt}}^{sur}$} & $H_{\text{no-tgt}}^{sur}$     & $\left\{\left(\boldsymbol{x}, y_{\text{tgt}}\right)\right\}  $          &                                                                                                &                                                                                                     &                                                                                         \\
MTA  \cite{guo2020multi}                   & \multicolumn{1}{c|}{$B, H_{\text{tgt}},H_{\text{no-tgt}}$}  & -                      & $\left\{\left(\boldsymbol{x}, y_{\text{tgt}},y_{\text{no-tgt}}\right)\right\}$ &                                                                                                &                                                                                                     &                                                                                         \\ \cline{4-7} 
FIA \cite{wang2021feature}                    & \multicolumn{1}{c|}{-}                        & $B,H_{\text{no-tgt}},H_{\text{tgt}}$ & $\left\{\left(\boldsymbol{x}, y_{tgt}\right)\right\}$            & \multirow{2}{*}{$\sqrt{ }$}                                                                    & \multirow{2}{*}{X}                                                                                  & \multirow{2}{*}{X}                                                                      \\
Practical no-box \cite{li2020practical}       & \multicolumn{1}{c|}{-}                        & $B,H_{\text{no-tgt}},H_{\text{tgt}}$ & $\left\{\left(\boldsymbol{x}, y_{\text{tgt}}\right)\right\} $           &                                                                                                &                                                                                                     &                                                                                         \\ \hline
HIT  \cite{zhang2022practical}                    & \multicolumn{1}{c|}{-}                        & $B,H_{\text{no-tgt}},H_{\text{tgt}}$ & $\left\{\left(\boldsymbol{x}\right)\right\}   $                  & \multirow{4}{*}{$\sqrt{}$}                                                                     & \multirow{4}{*}{$\sqrt{}$}                                                                          & \multirow{4}{*}{X}                                                                      \\
NRDM \cite{naseer2018task}                   & \multicolumn{1}{c|}{-}                        & $B,H_{\text{no-tgt}},H_{\text{tgt}}$ & $\left\{\left(\boldsymbol{x}\right)\right\}   $                  &                                                                                                &                                                                                                     &                                                                                         \\
DR  \cite{lu2020enhancing}                    & \multicolumn{1}{c|}{-}                        & $B,H_{\text{no-tgt}},H_{\text{tgt}}$ & $\left\{\left(\boldsymbol{x}\right)\right\} $                    &                                                                                                &                                                                                                     &                                                                                         \\
Cross-task+FGSM, PGD \cite{gurulingan2021uninet,sobh2021adversarial}   & \multicolumn{1}{c|}{$B, H_{\text{no-tgt}}$}          & $H_{\text{tgt}}$              & $\left\{\left(\boldsymbol{x}, y_{\text{no-tgt}}\right)\right\}$         &                                                                                                &                                                                                                     &                                                                                         \\ \hline
Ours                    & \multicolumn{1}{c|}{$B, H_{\text{no-tgt}}$}          & $H_{\text{tgt}}$              & $\left\{\left(\boldsymbol{x}, y_{\text{no-tgt}}\right)\right\}$         & $\sqrt{}$                                                                                      & $\sqrt{}$                                                                                           & $\sqrt{}$                                                                               \\ \hline
\end{tabular}}
\label{table: related attack}
\end{table*}

\textbf{Adversarial attack on multi-task classifiers.} Adversarial attacks against multi-task classifiers can be defined into two categories: single-task attacks and multi-task attacks \cite{mao2020multitask}. Single-task attacks aim to perturb the input $x$ such that the output of a specific target task $t_{\text{tgt}}$ differ from its ground-truth label $y_{\text{tgt}}$ (i.e., $F_{\text{tgt}}(\boldsymbol{x}+\boldsymbol{\delta}) \neq y_{\text{tgt}}$). To achieve this, optimization is performed to maximize $L_{i}$: 
\begin{equation}
\max _{\|\delta\|_p \leq \epsilon} \mathcal{L}_i\left(H_{\text {tgt }} \circ B(\boldsymbol{x}+\boldsymbol{\delta}), y_{\text{tgt}}\right)    
\end{equation}
where $H_{\text {tgt }}$ represents the head corresponding to the target task. The heads corresponding to non-target tasks are represented by $H_{\text {no-tgt }}$ in the following.

On the other hand, multi-task attacks aim to perturb the input $x$ such that the output of all tasks differs from their respective ground-truth labels (i.e.,$F_i(\boldsymbol{x}+\boldsymbol{\delta}) \neq y_i \quad \text { where } i \in\{1, \cdots,|T|\}$). To achieve this, optimization is performed to maximize the sum of losses for all tasks as follows:
\begin{equation}
    \max _{\|\boldsymbol{\delta}\|_p \leq \epsilon} \mathcal{L}_T(\boldsymbol{H}, B, \boldsymbol{x}+\boldsymbol{\delta}, \boldsymbol{y})
\end{equation}

Regarding adversarial attacks on multi-task models, \cite{gurulingan2021uninet,sobh2021adversarial} proposed to attack one of the multiple tasks. Due to the interrelation of tasks, they demonstrated that the effects of the attack on a single task affect other tasks. In this study, we refer to the attack method of indirectly attacking the target task through attacks on other tasks as cross-task attacks.

\subsection{Existing adversarial attack methods based on accessible knowledge}\label{sec:related attack}

In this subsection, we summarize the required access to the classification model and data when applying existing adversarial attack methods to multi-task classification models. Table \ref{table: related attack} summarizes the information required by the adversary to apply adversarial attacks to existing methods.

\textbf{Access to the target classifier:} In general, the adversary's access model to the target classifier can be classified into three levels based on the knowledge accessible to the attacker regarding the target classifier \cite{chen2017zoo}:
\begin{enumerate}
    \item White-box: the attacker has complete access to information such as architecture, parameters, outputs $F_i(x)$ for any input $x$, and so on.

    \item Black-box: the attacker can only observe outputs $F_i(\boldsymbol{x})$ for any input $\boldsymbol{x}$.

    \item No-box: the attacker has no access to any knowledge, including the output $F_{i}(\boldsymbol{x})$ for input $\boldsymbol{x}$.
\end{enumerate}

When the target is a multi-task classification model, access models of the head for each task and the backbone can be mutually different. The situation that we mainly consider is that the headers for non-target tasks and the backbone is accessible (e.g., white-box or black-box) while the header for the target tasks is not accessible (no-box).

\textbf{Access to the data:} In existing common white-box or black-box model attack methods \cite{goodfellow2014explaining,papernot2017practical}, it is assumed that labeled samples from both target and non-target tasks, obtained from the same distribution as the training data used for the model training, can be used. Existing methods for no-box attacks assume either the availability of labeled samples \cite{li2020practical} or unlabeled samples \cite{zhang2022practical,lu2023hard}.

\textbf{Attacker's capability:} Existing adversarial attack methods on multi-task classifiers can be summarized based on the knowledge accessible to the attacker, as presented in Table \ref{table: related attack}. 

First, we discuss the attack methods that require white or black box access to the header of the target task $H_{\text{tgt}}$.
As mentioned in \secref{sec:pre-attack}, FGSM and PGD (first row in Table \ref{table: related attack}) utilize gradients of the target classifier's loss function, requiring access to $H_{\text {tgt }} \circ B$. Thus, they cannot attack black-box classifiers directly. However, adversarial samples generated using these methods, by training a substitute classifier mimicking the target classifier's outputs and then attacking it with FGSM or PGD, have been shown effective against the target classifier as well \cite{papernot2017practical} (second row in Table \ref{table: related attack}). MTA \cite{guo2020multi} (third row in Table \ref{table: related attack}) proposes an attack method that generates adversarial perturbations for multiple tasks simultaneously. The key idea is to utilize a common feature encoder (similar to $B$ in Figure \ref{fig:multi-task}) to learn shared features among different tasks, then apply gradient-based attacks through this common feature encoder. They show that this strategy can more efficiently generate adversarial examples against multi-task models. The attacks mentioned are not applicable when the target task classifier is no-box.

Next, we discuss the attack methods that does not require any access to $H_{\text{tgt}}$. Attacks in the no-box setting fall into two categories. One is an attack that assumes that labeled data on the hidden target task is available and uses a substitute model for the target task trained on them. FIA \cite{wang2021feature} proposes attacks with strong transferability between architectures by generating a surrogate model trained with the target task's dataset and altering feature vectors in the direction that deteriorates the target task's loss. Similarly, \cite{li2020practical} proposes generating surrogate models using both supervised and unsupervised learning in situations where only a few labeled samples for the target task are available, transferring adversarial samples to the target. One limitation of these attack methods is that these attacks require labeled data for the target task to create a surrogate model for the target task and do not work when the labeled data of the target task is not available. Thus, these methodologies cannot be a solution to RQ1.

The other is to attack the model so that the classification performance on {\it arbitrary} tasks degrade.
HIT \cite{zhang2022practical} is a no-box attack method that uses a surrogate classifier generated by Contrastive Learning to generate adversarial examples without requiring labels of the target task. Similar to FIA, attacks focused on the backbone network, such as NRDM \cite{naseer2018task} and DR \cite{lu2020enhancing}, generate adversarial samples with high transferability to classifiers with various tasks or architectures. 

{Since we compare our proposal with NRDM and DR in the experimental section later, we introduce the attack algorithm of NRDM, DR, and cross task attack briefly.

{\bf Neural Representation Distortion Method (NRDM).} Let $B_k(\boldmath{x})$ be the feature map corresponds to the $k$th convolutional layer of the backbone network $B$. Then, the NRDM attack  generates adversarial perturbation $\delta^k_x$ by optimizing the following objective:
\begin{equation}
\underset{\left\|\boldsymbol{\delta}\right\|_p \leq \epsilon}{\operatorname{max}}||B_k(x+\delta)-B_k(x)||_2
\end{equation}
The NRDM uses optimization to find perturbations such that the feature map obtained from the perturbed input is maximally distant from the feature map of the true input.

{\bf Dispersion Reduction (DR).}
\label{backborn-attack}
Let $\mbox{std}$ be the function that evaluates the standard deviation over all pixels of a feature map. The DR attack optimizes the adversarial perturbation so that the standard deviation of all pixels of a specified feature map is maximized: 
  \begin{equation}
    \label{equation:dr}        \underset{\left\|\boldsymbol{\delta}\right\|_p \leq \epsilon}{\operatorname{max}} [-\mbox{std}(B_k(x+\delta))] \text { s.t. }\|\delta\|_p \leq \epsilon
    \end{equation}}

These attacks hypothesize that the shallow layers of the backbone network learn common features regardless of tasks or architectures. By generating adversarial samples that significantly alter the outputs (feature maps) of each layer of the surrogate classifier's backbone network, they achieve high transferability. Although these attacks are feasible even without any information on the target task, e.g., $y_{\text{tgt}}$, they lack stealthiness as they affect the performance of tasks other than the target task and cannot be a solution to RQ2.


{\bf Cross Task Attack.} \cite{chen2018multi} showed experimentally that in a multitask classifier, an attack on one task $t_j$ (attack task) may affect other tasks $t_i$ (target tasks). The following optimization obtains the adversarial perturbation generated by attacking $t_j$:
\begin{equation}
\label{eq:cross-task-attack}
   \underset{\left\|\boldsymbol{\delta}\right\|_p \leq \epsilon}{\operatorname{max}}\quad\mathcal{L}_j\left(x+\delta, y_j\right) 
\end{equation}
Cross-task attacks are relatively more effective when the behavior of the attack task and target task are highly similar.
Although cross-task attacks were not originally designed for the threat models discussed in this study, depending on the choice of attack task, it may be possible to generate attacks that affect the target task without access to the hidden target task. In practice, selecting the appropriate attack task for attacking the target task is difficult as the attacker does not know the target task. In our experiments, we attack all non-target tasks as attack tasks and evaluate the average as the attack effectiveness of a cross-task attack.

\section{Problem setup}
\label{sec:setup}
In this problem, we assume that the attacker knows that the multitask classifier was trained to predict tasks $t_{\text{no-tgt}}$ other than the target task $t_{\text{tgt}}$ ($t_{\text{no-tgt}} \in T \backslash t_{\text{tgt}}$). We generate an adversarial perturbation $\delta$ with $\|\boldsymbol{\delta}\|_p<\epsilon$ for sample $x$ such that when we input the adversarial sample $x+\delta$ to the target classifier, the prediction of $t_{\text{tgt}}$ by the multitask classifier will be different from the ground-truth and the prediction of $t_{\text{no-tgt}}$ will be the ground-truth. More specifically, we aim to generate small $\delta$ so that:

\begin{equation}
\Bigl\{\begin{array}{c}
F_{\mathrm{tgt}}(\boldsymbol{x}+\boldsymbol{\delta}) \neq y_{\mathrm{tgt}} \\
F_{\text {no-tgt }}(\boldsymbol{x}+\boldsymbol{\delta})=y_{\text {no-tgt }}
\end{array}
\label{eq:main}
\end{equation}

We then define a function to evaluate whether the above objectives are achieved for the task $t_i$. Let $D_T^{\text{adv}}=\left\{\left(\boldsymbol{x}^{\text{adv}}, \boldsymbol{y}\right)\right\}$ be the set of adversarial samples $\boldsymbol{x}^{\text{adv}}$ and ground-truth generated for each sample $(\boldsymbol{x}, \boldsymbol{y})$ in the dataset $D_T$ by the adversarial attack method. The effectiveness of the attack is determined by the difference between the accuracy of the model when clean samples are input, which is denoted as $\text { Accuracy }_i^{\text {clean }}$ and the accuracy of the model when adversarial samples are input, which is denoted as $\text { Accuracy }_i^{\text {adv }}$. More specifically, we define $\text { Accuracy }_i^{\text {clean }}$ and  $\text { Accuracy }_i^{\text {adv }}$ as follows: 
\begin{equation}
    \text { Accuracy }_i^{\text {clean }}=\frac{1}{\left|D_T\right|} \sum_{(\boldsymbol{x}, \boldsymbol{y}) \in D_T} \mathbf{1}\left[F_i(\boldsymbol{x})=y_i\right]\label{eq:c_acc}
\end{equation}
\begin{equation}
\text { Accuracy }_i^{\text {adv }}=\frac{1}{\left|D_T^{\mathrm{adv}}\right|} \sum_{\left(\boldsymbol{x}^{\mathrm{adv}}, \boldsymbol{y}\right) \in D_T^{\mathrm{adv}}} \mathbf{1}\left[F_i\left(\boldsymbol{x}^{\mathrm{adv}}\right)=y_i\right]\label{eq:a_acc}
\end{equation}

When $t_{i} = t_{\text{tgt}}$, $\text { Accuracy }_i^{\text {clean }} - \text { Accuracy }_i^{\text {adv }}$ evaluate the attack performance; otherwise, $\text { Accuracy }_i^{\text {clean }} - \text { Accuracy }_i^{\text {adv }}$ evaluate the stealthiness of non-targeted task. 

\subsection{Threat Model} \label{sec:threat}

\textbf{Access to the Target Classifier.} In existing studies, according to the definition in \secref{sec:threat}, classification as white/black/no-box is conducted only for the entire model. In other words, if $H_{\text{tgt}}$ is no-box, then $H_{\text{no-tgt}}$ and $B$ are also considered no-box. However, even if $H_{\text{tgt}}$ is no-box, in cases where the attacker has access to knowledge such as the parameters of pretrained models generally available or outputs for arbitrary inputs from $H_{\text{no-tgt}}$ or $B$, attacks may still be possible. Using such knowledge from $H_{\text{no-tgt}}$ or $B$, more effective attacks against $t_{\text{tgt}}$ than existing studies might be possible. 


With this in mind, we mainly consider the following situation in this study.
The attacker aims to attack a certain task, $t_{\text{tgt}}$, among the connected headers, assuming $H_{\text{tgt}}$ is no-box and cannot obtain knowledge about it. Conversely, the headers of tasks other than $t_{\text{tgt}}$, denoted as $H_{\text{no-tgt}}$. We consider two relationships between the attacker and $H_{\text{no-tgt}}$. First, $H_{\text{no-tgt}}$ are considered as white-box, meaning accessible to the attacker. Second, $H_{\text{no-tgt}}$ are considered as black-box, where only outputs are available. Additionally, $B$ is assumed to be a white box. This is justified when we can assume that a commonly known pre-trained model is used as the backbone, and this often happens in many realistic situations, expecially in image classification and language models.



\textbf{Access to Data.} Following \cite{lu2023hard}, we suppose the adversary can have any labeled or unlabeled data related to the target task is not available, while data related to  non-target tasks  are available.
In what follows, the labeled data for task groups $T \backslash t_{\text{tgt}}$ that the attacker can access is denoted by $D_{T \backslash t_{\text{tgt}}}$.

\textbf{Discussion on Attack Feasibility of Existing Methods in the Defined Threat Model.} 
 The feasibility of existing methods introduced in \secref{sec:threat} with the proposed threat model is discussed. 
In the threat model, access to $H_{\text{tgt}}$ is no-box, and $y_{\text{tgt}}$ is not obtained. Thus, attack methods that requre either of them are not available under this threat model.
For these reasons, existing attack methods that work in the white/black box setting, such as, FGSM, PGD, and MTA do not work in this threat model. 
Also, attack methods that rely on a substitute model of $H_{\text{tgt}} \circ B$, such as FIA and Practical no-box adversarial attack, are not available in the threat model because these methods require $y_{\text{tgt}}$ to train the substitute model.
NRDM, DR, cross-task attacks, and no-box attacks on single-task models such as HIT work in the threat model, as shown in Table \ref{table: related attack}. 
However, these attack methods other than the cross-task attack generate adversarial examples in a task-agnostic manner, which leads to degenerate the predictive performance of any task and lacks inherent stealthiness.
Leveraging specific information on the target task could lead to more effective attacks.

\section{Methodology}
\label{sec:methodology}
To achieve Eq. \ref{eq:main}, we propose an adversarial attack focusing on the features of neural networks that have been forgotten through catastrophic forgetting \cite{french1999catastrophic}.  

\subsection{Catastrophic forgetting attack}\label{sec: CF}
Catastrophic forgetting refers to the phenomenon where the knowledge of a task $t_i$ is lost after training a model with a different task $t_j$ \cite{french1999catastrophic}. For a multi-task classifier trained on a task set $T$, when fine-tuning is performed only on the subset of tasks $T \backslash t_{\mathrm{tgt}}$ excluding $t_{\text{tgt}}$, the tasks in $T \backslash t_{\mathrm{tgt}}$ maintain or improve their performance, while $t_{\text{tgt}}$ suffers from degraded performance due to catastrophic forgetting. Exploiting this property, we propose the Catastrophic Forgetting (CF) attack, which generates an adversarial example causes $t_{\text{tgt}}$ to make incorrect predictions while having minimal impact on the predictions of other tasks.

The forgetting of $t_{\text{tgt}}$ through catastrophic forgetting suggests that important features for predicting $t_{\text{tgt}}$ in the feature vector $B(\boldsymbol{x})$ may no longer be captured due to parameter updates of $B$. If we can reproduce the feature vector after catastrophic forgetting when adversarial examples are input, by adding small adversarial perturbations to the input, we can expect performance degradation in the target task. In other words, the CF attack causes performance degradation in the target task through catastrophic forgetting induced by retraining using data related to other tasks, even though the target task is not directly accessible to the attacker. 

CF attacks exploit the fact that target tasks are forgotten as a result of finetuning the model using only data about non-target tasks. In this sense, the CF attack does not need to utilize any information about the target task and works under our threat model. Furthermore, by utilizing the difference between the feature vectors before and after catastrophic forgetting, the CF attack can obtain an adversarial perturbation specifically defined for the the target task, even though it does not utilize any information about the target task.

\textbf{Algorithm} The details of the CF attack is summarized in Algorithm \ref{alg:main}. First the attacker aims to forget $t_{\text{tgt}}$, so fine-tuning is performed on $F_{T \backslash t_{\mathrm{tgt}}}$ with respect to $D_{T \backslash t_{\mathrm{tgt}}}$ (step 2). Let $FT$ be the function that returns the backbone network after fine-tuning, and $B^{\prime}=\mathrm{FT}\left(B, D_{T \backslash t_{\mathrm{tgt}}}\right)$ be the backbone network fine-tuned by $D_{T \backslash t_{\mathrm{tgt}}}$. Let $L_{CF}$ be the function that calculates the $L_{2}$ norm of $B(\boldsymbol{x}+\boldsymbol{\delta})$ and $B'(\boldsymbol{x})$ for a sample $\boldsymbol{x}$. Then, the CF attack is realized by the following approximate minimization of $L_{CF}$:
\begin{equation}
    \mathcal{L}_{C F}\left(B, B^{\prime}, \boldsymbol{x}+\boldsymbol{\delta}\right)=\left\|B(\boldsymbol{x}+\boldsymbol{\delta})-B^{\prime}(\boldsymbol{x})\right\|_2
\label{eq:l-cf}
\end{equation}
\begin{equation}
    \min _{\|\boldsymbol{\delta}\|_p \leq \epsilon} \mathcal{L}_{C F}\left(B, B^{\prime}, \boldsymbol{x}+\boldsymbol{\delta}\right)
\end{equation}

Furthermore, it is desirable for $B(\boldsymbol{x}+\boldsymbol{\delta})$ after approximation to behave the same as when input to the header of the target classifier as when input to the header after fine-tuning. Therefore, during fine-tuning, the parameters of the header are fixed.

In the algorithm \ref{alg:main}, we first fine-tune the model so that causes catastrophic forgetting in line 2. Then, from line 3 to line 8, we generate adversarial examples that realize our CF attack. More specifically, in lines 4-5, we apply a gradient-based method to generate perturbations that can close the distance between $B'(\boldsymbol{x})$ and $B(\boldsymbol{x}+\boldsymbol{\delta})$. In lines 6-7, we clip perturbation so that it satisfies the size constraint.

\begin{algorithm}
\caption{Catastrophic Forgetting Attack}\label{alg:main}
\begin{algorithmic}[1]
\Require: Dataset $D_{T \backslash t_{\text{tgt}}}$, clean sample $x$, backbone network of target model $B$, the number of iteration $I$, step size $\alpha$, max perturbation $\epsilon$
\State \texttt{Initialize: $\boldsymbol{x}^{\text{adv}} \leftarrow \boldsymbol{x}$}
\State \texttt{$B^{\prime}=\text{FT}\left(B, D_{T \backslash t_{\text{tgt}}}\right)$}
\For{i=0,...,I}
    \State \texttt{$\boldsymbol{g}=\nabla_{\boldsymbol{x}^{\mathrm{adv}}} L_{C F}\left(B, B^{\prime}, \boldsymbol{x}^{\text{adv}}\right)$}
    \State \texttt{$\boldsymbol{x}^{\text{adv}}=\boldsymbol{x}^{\mathrm{adv}}+\alpha \cdot \operatorname{sign}(\boldsymbol{g})$}
    \State \texttt{$\boldsymbol{x}^{\text{adv}}=\boldsymbol{x}+\operatorname{clip}_{-\epsilon,\epsilon}\left(\boldsymbol{x}^{\text{adv}}-\boldsymbol{x}\right)$}  \State \texttt{$\boldsymbol{x}^{\text{adv}}=\operatorname{clip}_{0,255}\left(\boldsymbol{x}^{\text{adv}}\right)$}   
\EndFor
\State{return $x^{\text{adv}}$}    
\end{algorithmic}
\end{algorithm}

\subsection{CF delta attack} \label{sec: CFdelta}
In the last subsection, we proposed a CF attack to replicate the feature vectors after catastrophic forgetting. Since the degradation of the target task's accuracy in CF attack depends on the degree of degradation in the overall classification performance after forgetting, the attack performance significantly decreases for tasks that are less prone to forgetting. Therefore, to further degrade the accuracy of the target task, we propose an attack called CF delta attack, which focuses not only on replicating the feature vectors after forgetting but also on the change in the direction of feature vectors before and after forgetting. This idea is based on the hypothesis that even for tasks less prone to forgetting, the feature vectors after fine-tuning tend to change in the direction that induces forgetting.


The CF delta attack significantly changes the feature vectors compared to the CF attack, which may lead to a greater impact on tasks other than the target task, thereby compromising stealthiness. Therefore, we expect to maintain stealthiness by adding a penalty term to preserve the performance of other tasks.

\textbf{Algorithm:} For the CF delta attack, but we replace $L_{CF}$ with a new loss function $L_{CF_{\Delta}}$ as defined in Equation \ref{eq:cf_Delta}.

We denote the function that returns the difference of feature vectors before and after fine-tuning as $\Delta$:
\begin{equation}
    \Delta\left(B, B^{\prime}, \boldsymbol{x}\right)=B^{\prime}(\boldsymbol{x})-B(\boldsymbol{x})
\end{equation}

Intuitively, we expect that, for tasks with weak forgetting effects, the forgetting effect can be strengthened by shifting the post-forgetting feature vector $B'(\boldsymbol{x})$ further towards the direction of $\Delta\left(B, B^{\prime}, \boldsymbol{x}\right)$, the difference between before and after forgetting. This idea can be formulated as a loss function as follows
\begin{equation}
    L_{CF_{\Delta}}(B, \Delta, \beta, \boldsymbol{x})=\|B(x+\delta)-(B(x)+\beta \Delta)\|_2 \label{eq:cf_Delta}
\end{equation}
where $\beta$ is a hyperparameter that controls the effect of $\Delta$.

To further ensure stealthiness, we optimize $\boldsymbol{\delta}$ by adding a penalty term to maintain the performance of the non-target tasks:
\begin{equation}
    \min _{\|\boldsymbol{\delta}\|_p \leq \epsilon} L_{C F_{\Delta}}(B, \Delta, \beta, \boldsymbol{x})+\gamma \mathcal{L}_{T \backslash t_{\text {tgt }}}(\mathbf{H}, B , \boldsymbol{x}, \boldsymbol{y}) \label{eq:cf_delta_obj}
\end{equation}
$\beta$ and $\gamma$ are hyperparameters.
Greater $\beta$ is expected to strengthen the attack performance while degrading stealthiness.
Greater $\gamma$ is expected to improve the stealthiness while weakens the attack performance.
When adjusting $\beta$, since the attacker cannot directly evaluate the attack performance on $t_{\text{tgt}}$, the optial $\beta$ and $\gamma$ parameters are selected such that the degradation in performance of $t_{\text{no-tgt}}$ does not exceed a certain threshold.

\section{Experiments}
\label{sec:eval}


In this section, we show experimental results to verify whether the adversarial samples generated using the proposed method satisfy the attacker's objective, (Equation \ref{eq:main}. Specifically, we generate adversarial samples using the proposed method and existing methods discussed in \secref{sec:threat}, then measure $\text{ Accuracy}_i^{\text {adv}}$ for each task $t_i$, and compare it with $\text{ Accuracy}_i^{\text {clean}}$. Additionally, we compare the effectiveness of the attack (1) when the target classifier's $H_{\text {no-tgt }}$ is black box and a white box, and  (2) when the task headers $H_t$ are linear layers and nonlinear layers.

The goal of the proposed method is to produce incorrect predictions for $t_{\text{tgt}}$ (attack on the hidden task, RQ1) while maintaining correct predictions for tasks other than $t_{\text{tgt}}$ (stealthiness for non-hidden tasks). Therefore, we expect $\text{ Accuracy}_{\text{tgt}}^{\text {adv}}$ - $\text{ Accuracy}_{\text{tgt}}^{\text {clean}}$  to be large and $\text{ Accuracy}_{\text{no-tgt}}^{\text {adv}}$ - $\text{ Accuracy}_{\text{no-tgt}}^{\text {clean}}$ to be small compared to existing methods.

\subsection{Experimental Settings}
\textbf{Dataset:} We use the CelebA \cite{celeb} and DeepFashion \cite{deepfashion} datasets in our experiments. In both CelebA and DeepFashion datasets, each sample contains multiple attributes.
The attributes and labels adopted in this study are shown in Table \ref{table:attribute}. 
We can regard the prediction of these attributes as tasks in the multi-task learning setting.  
That is, we defined CelebA as a four-task problem and DeepFashion as a three-task problem.
Let $D_T$ be the dataset. 10\% of $D_T$ was used by the attacker for fine-tuning and generating adversarial samples. The remaining data was used for training the target classifier.

\begin{table}[ht]
\caption{Attributes of CelebA and DeepFashion datasets}
\begin{tabular}{c|c|c}
Dataset                      & Attribute (Abbreviation) & Label                 \\ \hline
\multirow{5}{*}{CelebA}      & Male(u)                 & yes,no                \\
                             & Wavy Hair(H)            & yes,no                \\
                             & Wearing Lipstick(K)     & yes,no                \\
                             & Smiling(F)              & yes,no                \\
                             & Black Hair(i)           & yes,no                \\ \hline
\multirow{3}{*}{DeepFashion} & Category(C)             & upper,lower,full      \\
                             & Sleeve(S)               & long,short,sleeveless \\
                             & Neckline(N)             & v,crew,square,no      \\ \hline
\end{tabular}
\label{table:attribute}
\end{table}

\textbf{Target classifier:} The target classifier architecture consists of a ResNet-18 backbone network $B$ and one linear layer for each task header $H_{\mbox{tgt}}$ and $H_{\mbox{no-tgt}}$. The weights $\lambda_{i}$ in the loss function (Equation \ref{eq:multi-losses}) are all set to 1.

\textbf{Comparison methods:} Comparison methods include the cross-task attack, NRDM, and DR attacks, which are attack methods that work in the threat model. The original NRDM and DR methods employ substitution models that simulate each layer of the backbone network, and the adversarial perturbation is generated so that the post-perturbation feature vector is as far away as possible from the pre-perturbation feature vector. In our threat model, the attacker is allowed to access the backbone network of the target classifier directly. Therefore, NRDM and DR attacks are conducted not with substitute models but with the true backbone network.
In the cross-task attack, the adversary chooses a non-hidden task as a proxy and generates adversarial samples for the selected non-target task, expecting that the attack takes effect if the selected task is strongly correlating with the hidden target task. In practice, it is difficult for the attacker to select an appropriate proxy task as it has no knowledge of the hidden target task. In the experiments, the cross-task attack is performed on all non-target tasks and the average performance of all tasks is shown as the result. 

\textbf{Hyperparameter:} Following the hyperparameter tuning study \cite{mao2020multitask}, experiments are conducted with $\epsilon = 8$ for the attack. 
$\beta$ and $\gamma$ in the CF delta attack is selected such that the degradation in performance of $t_{\text{no-tgt}}$ does not exceed a certain threshold.
More specifically, parameters were selected to ensure that the accuracy of $t_{\text{no-tgt}}$ did not degrade by more than 0.1 in comparison to clean accuracy. Specific parameter settings are indicated in the legend of the result figures. Detailed discussions on the hyperparameter tuning are shown in Section \ref{sec:hyper}.

\textbf{Threat model:} $B$ is considered as white-box, $H_{\text{tgt}}$ is considered as no-box. We conduct experiments with $H_{\text{no-tgt}}$ in both white-box and black-box setting.

\subsection{Results}

\begin{table*}[th]
\caption{Results of the attack performance/stealthiness against the target task/non-target task in CelebA and Deepfashion datasets under \textbf{white-box setting with linear task head}. Here, the former value represents attack performance, which was evaluated with classification accuracy on the target task $(\%)$. Lower means better attack performance; the latter value represents stealthiness, which was evaluated with classification accuracy on the non-target task $(\%)$. Higher means better stealthiness}
\scalebox{0.9}{\begin{tabular}{c|ccccc|ccc}
\multicolumn{1}{c|}{}         & \multicolumn{5}{c|}{CelebA}                                                                                           & \multicolumn{3}{c}{DeepFashion}                                    \\ \hline
        & u                     & H                     & K                     & F                     & i                     & C                    & S                    & N                    \\ \hline
Clean                         & 97.66/97.66           & 81.20/81.20           & 93.02/93.02           & 92.11/92.11           & 88.18/88.18           & 81.83/81.83          & 85.70/85.70          & 70.88/70.88          \\
NRDM                          & 63.46/63.46           & 56.45/56.45           & 60.30/60.30           & 73.84/73.84           & 66.14/66.14           & 41.50/41.50          & 44.30/44.30          & 40.10/40.10          \\
DR                            & 59.27/59.27           & 67.29/67.29           & 48.35/48.35           & 49.18/49.18           & 75.98/75.98           & 19.60/19.60          & 52.25/52.25          & 42.60/42.60          \\
Cross-task                    & 62.81/62.81           & 59.21/59.21           & 51.72/51.72           & 68.34/68.34           & 76.35/76.35           & 52.15/52.15          & 48.9/48.9            & 43.92/43.92          \\
CF                            & 89.16/98.24           & 69.28/86.03           & 88.26/93.72           & 49.17/93.59           & 88.95/89.50           & 49.20/91.05          & 62.90/\textbf{99.63} & 64.65/95.44          \\
CF delta($\beta=10,\gamma=0$) & 40.22/94.63           & 49.98/94.40           & 46.04/88.12           & 36.81/94.67           & 72.20/96.25           & \textbf{16.00}/36.16 & 18.93/42.475         & \textbf{23.58}/66.54 \\
CF delta($\beta=20,\gamma=1$) & \textbf{37.44/99.76} & \textbf{23.73/99.80} & \textbf{38.57/99.82} & \textbf{30.59/99.82} & \textbf{41.66/99.68} & 18.88/\textbf{93.15} & \textbf{17.70}/96.00 & 29.33/\textbf{97.28} \\ \hline
\end{tabular}}
\label{table:white-box}
\end{table*}
   
\subsubsection{White-box Setting}

We first consider the case where the adversary has white-box access to $H_{\text{no-tgt}}$, the header for nontarget tasks.
When $H_{\text{no-tgt}}$ consists of linear layers, the attack performance and stealthiness against the target task of proposed methods and comparison methods for CelebA and Deepfashion are shown in Table \ref{table:white-box}. In Table \ref{table:white-box}, the third row shows the test accuracy on the clean data; the fourth, fifth, and sixth rows show attack performance and stealthiness of NRDM, DR, and Cross-task attack. Since these three attacks do specify a target task when generating adversarial examples against a hidden target task, the attack performance and stealthiness have the same value for different target tasks. We can observe although these attacks have good attack performance, they do not have good stealthiness. This is because these attacks aim to generate adversarial examples with good transferability, hence, they attack different tasks at the same time. The seventh, eighth, and ninth rows in Table \ref{table:white-box} show attack performance and stealthiness of the proposed CF attack and CF delta attack with different $\beta$ and $\gamma$. We can observe our proposal achieves not only good attack performance compared with the existing attack but also good stealthiness. Achieving good stealthiness is a property never achieved by previous methods.

When $H_{\text{no-tgt}}$ consists of non-linear layers, the attack performance and stealthiness against the target task of proposed methods and comparison methods for CelebA and Deepfashion are shown in Table \ref{table:white-box-nolinear}. We can observe a similar tendency that our proposal achieves not only good attack performance but also good stealthiness. These results show that our proposal can also work with non-linear head layers.

\subsubsection{Black-box Setting}

We next consider the case where the adversary hasonly  black-box access to $H_{\text{no-tgt}}$.
When $H_{\text{no-tgt}}$ is a black box and task heads are linear layers, the attack performance and stealthiness against the target task of proposed methods and comparison methods for CelebA and Deepfashion are shown in Table \ref{table:black-box-linear}. Since, NRDM and DR are not affected by $H_{\text{no-tgt}}$, hence fourth and fifth rows in Table \ref{table:black-box-linear} are the same as fourth and fifth rows in Table \ref{table:white-box}. Our proposals still achieve the best attack performance and stealthiness in the black-box setting. By comparing Table \ref{table:black-box-linear} and Table \ref{table:white-box-nolinear}, we can observe that in the black box setting, the attack performance and stealthiness are worse than those in the white box setting. However, the proposed methods still work well in the black box setting, and degrade model's performance on target task without affecting the non-target tasks a lot.

Among Table \ref{table:white-box}, \ref{table:white-box-nolinear}, and \ref{table:black-box-linear}, we can also observe that CF attack is not always effective, its performance largely depends on the target task. For example, on CelebA dataset, CF attack works well on (F) task on all settings (fifth column and seventh row in Table \ref{table:white-box}, Table \ref{table:white-box-nolinear}, and Table \ref{table:black-box-linear}). However, CF attack does not degrade the model performance a lot on task (K) task (fifth column and seventh row in Table \ref{table:white-box}, Table \ref{table:white-box-nolinear}, and Table \ref{table:black-box-linear}). In contrast, CF delta attacks perform well on all target tasks. However, CF delta attack does not always have good stealthiness when $\gamma$ and $\beta$ are not set up well. For example, in Table \ref{table:white-box}, CF delta attack shows limited stealthiness on DeepFashion dataset on task C when $\gamma=0$ (seventh row and seventh column). When $\gamma=1$, the stealthiness can be largely improved without affecting attack performance. This shows the design of the penalty term is important for stealthiness.

\begin{table*}[th]
\caption{Results of the attack performance/stealthiness against the target task/non-target task in CelebA and Deepfashion datasets under \textbf{white-box setting with non-linear task head}. Here, the former value represents attack performance, which was evaluated with classification accuracy on the target task $(\%)$. Lower means better attack performance; the latter value represents stealthiness, which was evaluated with classification accuracy on the non-target task $(\%)$. Higher means better stealthiness}
\scalebox{0.9}{\begin{tabular}{c|ccccc|ccc}
\multicolumn{1}{l|}{}         & \multicolumn{5}{c|}{CelebA}                                                                                      & \multicolumn{3}{c}{DeepFashion}                                    \\ \hline
Target/Non-target task        & u                    & H                    & K                    & F                    & i                    & C                    & S                    & N                    \\ \hline
Clean                         & 97.79/97.79          & 81.43/81.43          & 93.09/93.09          & 92.01/92.01          & 87.95/87.95          & 77.72/77.72          & 82.15/82.15          & 66.55/66.55          \\
NRDM                          & 60.64/60.64          & 50.31/50.31          & 51.78/51.78          & 67.76/67.76          & 67.71/67.71          & 46.75/46.75          & 37.05/37.05          & 28.88/28.88          \\
DR                            & 61.45/61,45          & 42.30/42.30          & 52.15/52.15          & 50.03/50.03          & 72.85/72.85          & 19.50/19.50          & 51.78/51.78          & 42.15/42.15          \\
Cross-task                    & 73.14/73.14          & 58.15/58.15          & 65.63/65.63          & 73.53/73.53          & 74.57/74.47          & 44.93/44.93          & 46.86/46.86          & \textbf{28.39}/28.39 \\
CF                            & 90.53/\textbf{99.98} & 68.19/\textbf{99.98} & 90.32/\textbf{99.98} & 61.24/\textbf{99.99} & 72.84/\textbf{99.99} & 63.53/\textbf{97.39} & 75.70/\textbf{99.06} & 63.55/\textbf{97.14} \\
CF delta($\beta=10,\gamma=0$) & 40.78/99.49          & 26.89/85.65          & 56.52/92.52          & 20.84/86.86          & 48.89/94.00          & 29.55/85.71          & 60.68/92.28          & 49.48/88.66          \\
CF delta($\beta=20,\gamma=1$) & \textbf{28.86}/99.61 & \textbf{20.84}/82.96 & \textbf{21.83}/94.98 & \textbf{11.39}/89.24 & \textbf{32.53}/93.51 & \textbf{18.95}/67.92 & \textbf{36.08}/70.31 & 36.85/72.42          \\ \hline
\end{tabular}}
\label{table:white-box-nolinear}
\end{table*}

\begin{table*}[th]
\caption{Results of the attack performance/stealthiness against the target task/non-target task in CelebA and Deepfashion datasets under \textbf{blcak-box setting with linear task head}. Here, the former value represents attack performance, which was evaluated with classification accuracy on the target task $(\%)$. Lower means better attack performance; the latter value represents stealthiness, which was evaluated with classification accuracy on the non-target task $(\%)$. Higher means better stealthiness}
\scalebox{0.9}{\begin{tabular}{c|ccccc|ccc}
\multicolumn{1}{l|}{}         & \multicolumn{5}{c|}{CelebA}                                                                                      & \multicolumn{3}{c}{DeepFashion}                                    \\ \hline
Target/Non-target task        & u                    & H                    & K                    & F                    & i                    & C                    & S                    & N                    \\ \hline
Clean                         & 97.66/97.66           & 81.20/81.20           & 93.02/93.02           & 92.11/92.11           & 88.18/88.18           & 81.83/81.83          & 85.70/85.70          & 70.88/70.88          \\
NRDM                          & 63.46/63.46           & 56.45/56.45           & 60.30/60.30           & 73.84/73.84           & 66.14/66.14           & 41.50/41.50          & 44.30/44.30          & 40.10/40.10          \\
DR                            & 59.27/59.27           & 67.29/67.29           & 48.35/48.35           & 49.18/49.18           & 75.98/75.98           & 19.60/19.60          & 52.25/52.25          & 42.60/42.60          \\
Cross-task                    & 64.18/64.18           & 56.67/56.67           & 50.92/50.92          & 67.94/67.94           & 74.10/74.10          & 48.5/48.5        & 46.65/46.65            & 38.64/38.64          \\
CF                            & 90.86/99.88          & 65.80/93.14          & 90.29/98.76          & 50.10/98.97          & 72.84/97.01          & 41.95/83.18          & 60.42/\textbf{99.84} & 64.18/92.08          \\
CF delta($\beta=10,\gamma=0$) & 68.13/98.69          & 50.85/88.17          & 75.86/86.42          & 35.54/97.07          & 72.35/94.42          & 20.87/67.48          & 18.63/69.11          & 32.22/74.41          \\
CF delta($\beta=20,\gamma=1$) & \textbf{49.54/99.98} & \textbf{35.62/99.81} & \textbf{46.69/99.87} & \textbf{22.89/99.98} & \textbf{56.93/99.65} & \textbf{17.00/91.61} & \textbf{16.92}/93.48 & \textbf{28.92/95.53} \\ \hline
\end{tabular}}
\label{table:black-box-linear}
\end{table*}

\section{Discussion}

\begin{figure*}
    \centering
    \includegraphics[width=15cm]{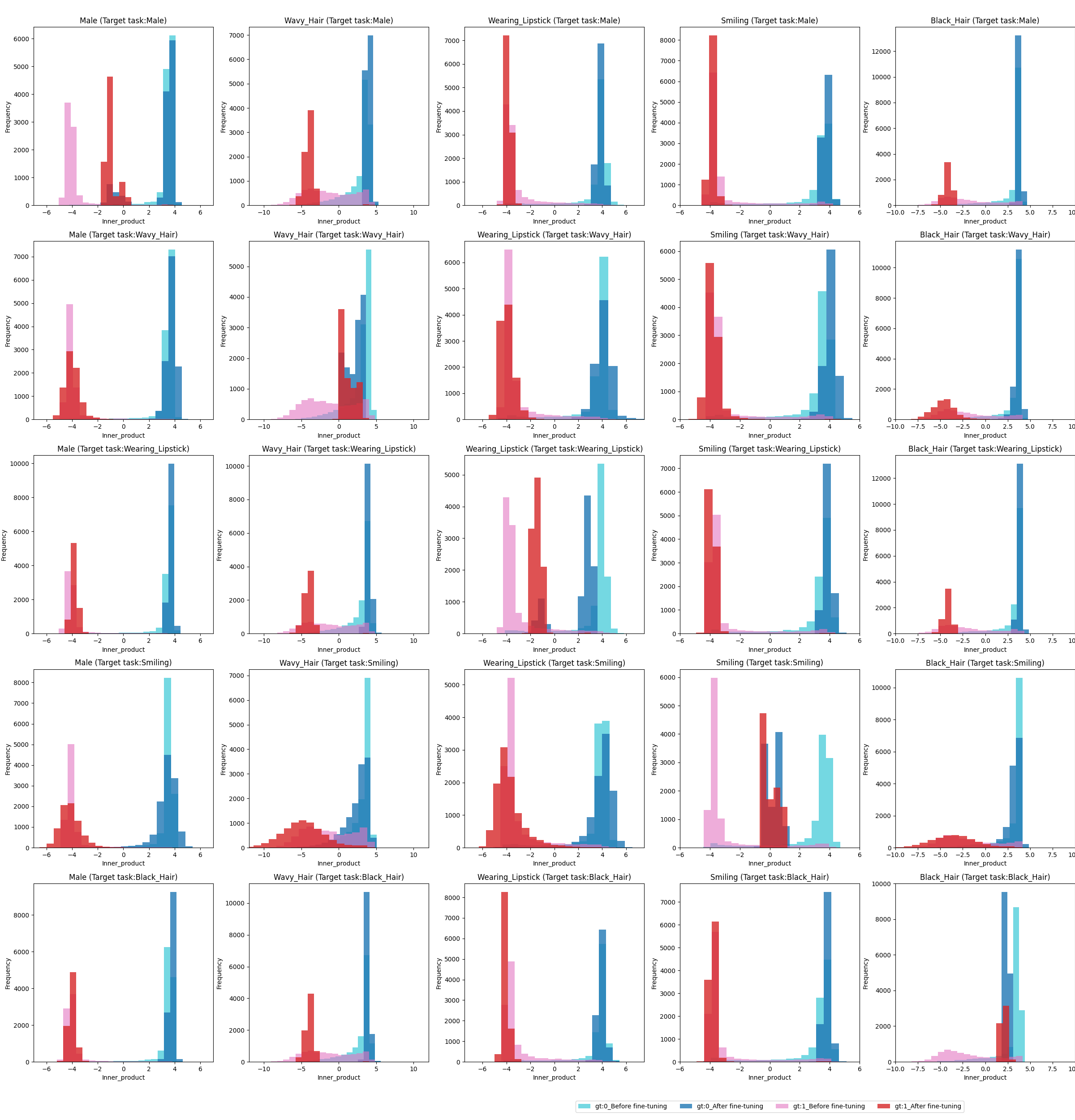}
    \caption{Distribution of the inner product of the feature vector before and after catastrophic forgetting and the weight vector of the header of each task in CelebA.  The larger this inner product, the greater the predictive probability of predicting the label 1(yes). Each row represents the target task, and each column represents the header of the task used to compute the inner product. The distribution of the inner product for the target task (diagonal elements) has its center away from zero before fine-tuning but after fine-tuning, its center shifts towards zero. This indicates that constructing the adversarial example using the post-fine-tuning model may degrade the target task's classification performance. On the other hand, the distribution of the inner product (off-diagonal elements) in the non-target task does not change much with fine-tuning, which means that fine-tuning has little impact on the non-target task, which is expected to attain highly stealthy behavior when used in attacks.}
    \label{fig:feature-vector-and-weight-inner-product}
\end{figure*}

\subsection{Effect of Catastrophic Forgetting}
The proposed method assumes that features useful for the target task that are remembered before finetuning are lost after finetuning, and the adversary exploits the differences in feature vectors before and after catastrophic forgetting for the attack. Experimental considerations were conducted to check whether this assumption is valid in attacks.

We considered the simplest case, in which the header consists of a linear layer for each task (Makeup, Wavy hair, Wearing lipstick, Smiling, Black hair) in Celeb A, where each task is binary classified.

Let $B(\boldmath{x})$ be an input, $\boldmath{z}=B(\boldmath{x})$ be the input to the header consisting of linear layers, and $\boldmath{w}_k$ be the weight vector corresponding to the class $k$ of the linear layer. In the binary classification problem, the larger $\boldmath{w}_1^T\boldmath{z}$ is, the higher the probability that $\boldmath{z}$ is predicted to be class $1$. 
Therefore, in well-trained neural networks, $\boldmath{w}_1^T\boldmath{z}$ is expected to distribute around some positive values for feature vectors with ground truth 1 and around some negative values for feature vectors with ground truth 0. 
On the other hand, if catastrophic forgetting happens for the task, the two classes will no longer be distinguishable, and it is expected that the separability of the two distributions will diminish; the centers of both distributions will approach zero.

\begin{figure*}[th]
    \centering
    \includegraphics[width=\linewidth]{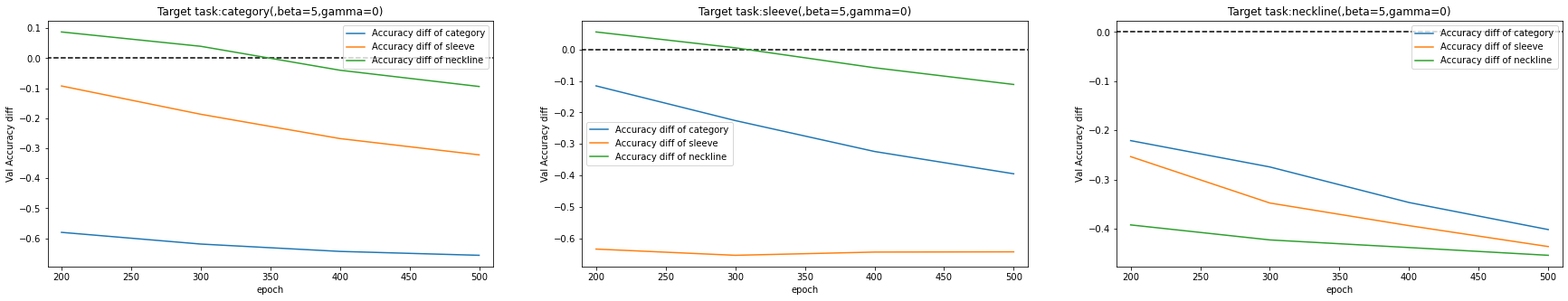}
    \caption{Number of fine-tuning epochs (horizontal axis) vs Adv Accuracy against Clean Accuracy $\text{Accuracy}^{\text{adv}}_{\text{tgt}}- \text{Accuracy}^{\text{clean}}_{\ text{tgt}}$ (vertical axis) in the three tasks of DeepFashion (lower is larger attack effect for the target task and less stealthiness for the non-target tasks).  The target task was set as Category, Sleeve, and Neckline, from left to right.}
    \label{fig:CF_delta_vs_epoch}
\end{figure*}


\begin{figure*}[th]
    \centering
    \includegraphics[width=\linewidth]{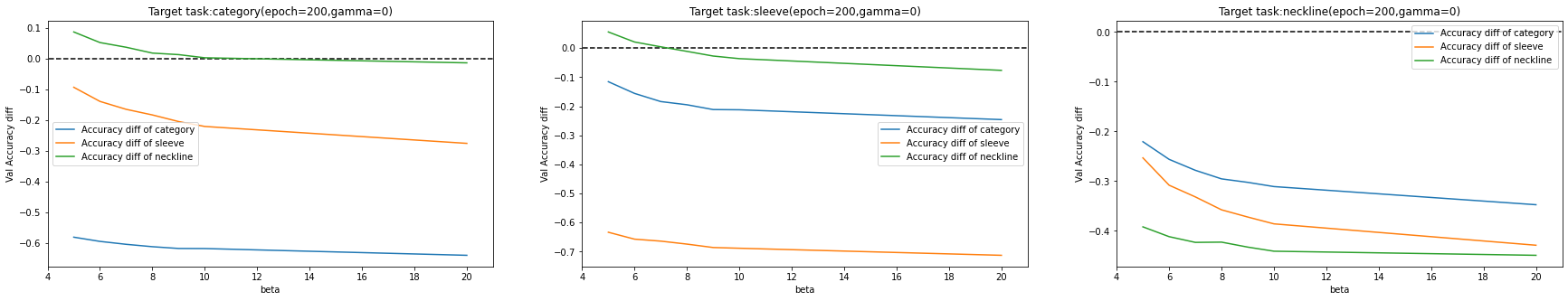}
    \caption{$\beta$ (the extent to which the change in the feature vector before and after fine-tuning is magnified, horizontal axis) vs Adv Accuracy against Clean Accuracy $\text{Accuracy}^{\text{adv}}_{\text{tgt}}- \text{Accuracy}^{\text{clean}}_{\ text{tgt}}$ (vertical axis) in the three tasks of DeepFashion (lower is larger attack effect for the target task and less stealthiness for the non-target tasks).  The target task was set as Category, Sleeve, and Neckline, from left to right.}
    \label{fig:CF_delta_vs_beta}
\end{figure*}


\begin{figure*}[th]
    \centering
    \includegraphics[width=\linewidth]{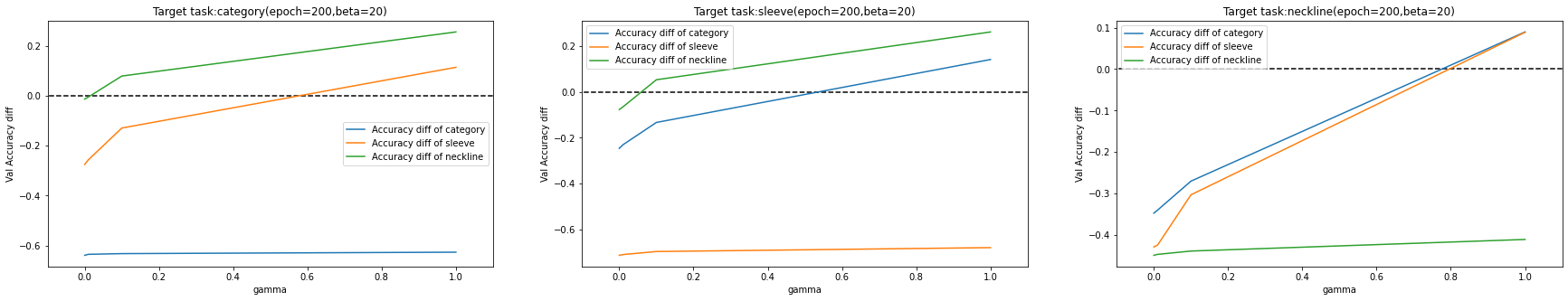}
    \caption{$\gamma$ ( a parameter that balances the promotion of stealthness and strength of attack, horizontal axis) vs Adv Accuracy against Clean Accuracy $\text{Accuracy}^{\text{adv}}_{\text{tgt}}- \text{Accuracy}^{\text{clean}}_{\ text{tgt}}$ (vertical axis) in the three tasks of DeepFashion (lower is larger attack effect for the target task and less stealthiness for the non-target tasks).  The target task was set as Category, Sleeve, and Neckline, from left to right.}
    \label{fig:CF_delta_vs_beta}
\end{figure*}

In the experiments, one of the five tasks in CelebA (target task) was chosen as the target of catastrophic forgetting, and the changes in the inner product distributions of feature vectors before and after forgetting are evaluated. The results are shown in Figure\ref{fig:feature-vector-and-weight-inner-product}. 
Each row represents the task targeted for catastrophic forgetting, and each column represents the task for which the inner product was evaluated.
When the target task and the task for which the inner product was evaluated coincide (figures located in the diagonal components in Figure \ref{fig:feature-vector-and-weight-inner-product}, the pre-forgetting distributions (light blue and pink ) are expected to separate well, while the post-forgetting distribution (blue and red) is expected to shift around 0, worsening separability.
On the other hand, if the target task and the task for which the inner product was evaluated do not coincide (figures located in the non-diagonal components in Figure \ref{fig:feature-vector-and-weight-inner-product}), the separability is not expected to change before and after forgetting. 
Figure \ref{fig:feature-vector-and-weight-inner-product} shows that the experimental results generally behave as expected above. This result suggests that attacks that utilize differences before and after catastrophic forgetting can be stealthy, with little impact on non-target tasks.
Interestingly, the degree of change in the distribution varies across tasks. For instance, the task of "Wearing Lipstick" shows less change compared to "Smiling." This variability in forgetting difficulty suggests that CF attacks may be relatively less effective in certain tasks. However, the center of the distributions also shifts toward zero in these tasks.  This trend may provide evidence that, with appropriate parameter tuning, CF delta attacks show better attack performance than CF attacks.

\subsection{Accuracy Improvement of Non-target Tasks After CF/CF-delta Attack}

When the adversary mounts attack the model using the CF and CF-delta attacks, the accuracies of the non-target tasks after the attack were often improved in Table \ref{table:white-box}, \ref{table:white-box-nolinear}, and \ref{table:black-box-linear}. In this subsection, we discuss the reasons for this. 

If the CF and CF delta attacks work as expected, then $B(\boldsymbol{x}^{\text{adv}}) \simeq B^\prime(\boldsymbol{x})$ holds where  $\boldsymbol{x}^{\text{adv}}$ is the adversarial example generated by the attacker, $B$ is the backbone of the target model and $B^\prime$ is the backbone after finetuning by the attacker. If $B(\boldsymbol{x}^{\text{adv}})$ is used for the target task, finetuning can cause catastrophic forgetting, which is more likely to cause performance degradation. 
On the other hand, if $B(\boldsymbol{x}^{\text{adv}})$ is used for the non-target tasks, the classification performance can be improved because the model is finetuned with data that has never been used for the model training by the model owner. 
In the experiments, the model owner used 90\% of data for training and the attacker 10\% of data for finetuning (e.g., CF/CF-delta attack). For this, an improvement in the accuracy of the non-target task was considered to be observed.

\subsection{Hyperparameter Tuning without using Target Task}\label{sec:hyper}

The hyperparameter tuning strategies for the CF and CF delta attacks are described: for the CF attack, the number of epochs for which fine-tuning is performed during the attack is a hyperparameter; for the CF delta attack, in addition to this, the $\beta$ shown in eq. \ref{eq:cf_Delta} $\beta$ and $\gamma$ shown in eq. \ref{eq:cf_delta_obj} shall be hyperparameters. The $\beta$ is a parameter that determines the extent to which the change in the feature vector before and after fine-tuning is magnified. 
The $\gamma$ is a parameter that balances the promotion of stealthness and strength of attack: larger epoch numbers and $\beta$ are expected to increase the effectiveness of the attack but may impair stealth if excessively large.

In the threat models we consider in this study, attackers cannot have any information about the target task. Therefore, tuning the hyperparameters using some indicators related to the target task, such as the success rate of attacks, is impossible. Therefore, hyper-parameter tuning must be carried out using information related to the non-target tasks.
For this reason, we employed the change in the accuracy before and after an attack on the non-target tasks as the indicator, which the attacker can observe. Specifically, we adopt the largest parameter within which accuracy does not deteriorate by more than 0.1 after the attack in any non-target task. 

For three of tasks deepfashion (Category, Sleeve, Neckline), we evaluarted the difference of clean accuracy $\text{Accuracy}^{\text{adv}}_{\text{tgt}}$ (eq. \ref{eq:c_acc}) and adversarial accuracy $\text{Accuracy}^{\text{adv}}_{\text{tgt}}$ (eq. \ref{eq:a_acc})  when varying the three hyper-parameters  were experimentally evaluated.

Figures \ref{fig:CF_delta_vs_epoch}, \ref{fig:CF_delta_vs_beta}, and \ref{fig:CF_delta_vs_beta} demonstrate the changes of $\text{Accuracy}^{\text{adv}}_{\text{tgt}}- \text{Accuracy}^{\text{clean}}_{\text{tgt}}$ (horizontal axis) when the number of fine-tuning epochs, $\beta$ and $\gamma$ (vertical axis) were varied for the three tasks, Category, Sleeve, and Neckline, respecitvely.
In all figures, the target tasks were set from left to right as Category, Sleeve and Neckline.

The smaller $\text{Accuracy}^{\text{adv}}_{\text{tgt}}- \text{Accuracy}^{\text{clean}}_{\text{tgt}}$ is in the negative direction, the greater the performance degradation due to the attack.
Therefore, it is desirable for the attacker that this difference is large in the negative direction for the target task (e.g, large attack effect) and small for the non-target tasks (e.g., strong stealthiness).

From the results in Figures  \ref{fig:CF_delta_vs_epoch} - \ref{fig:CF_delta_vs_beta}, we can see that the attack effect on the target task is relatively larger than that on the non-target task, indicating a certain degree of success in maintaining stealthiness.
It can be seen that an increase in $\beta$ and $\gamma$ causes an increase in attack effectiveness for both target and non-target tasks.
On the other hand, an increase in $\gamma$ has little effect on the attack effect on the target task and causes a decrease in the attack effect on the non-target task. This indicates that regularization by  $\mathcal{L}_{T \backslash t_{\text {tgt }}}(\mathbf{H}, B , \boldsymbol{x}, \boldsymbol{y})$ in eq. \ref{eq:cf_Delta} of works usefully in increasing stealthiness.
We remark that the attacker cannot use the attack effect on the target task $\text{Accuracy}^{\text{adv}}_{\text{tgt}}- \text{Accuracy}^{\text{clean}}_{\text{tgt}}$
to tune these parameters.
As it can be confirmed that the attack effect on the target task is proportional to the attack effect on the non-target task empirically, we selected hyperparameters to the extent that $\text{Accuracy}^{\text{adv}}_{\text{no-tgt}}- \text{Accuracy}^{\text{clean}}_{\text{no-tgt}}$ for a single task did not fall below -0.1. 

\section{Conclusion}
\label{sec:conclusion}
This study proposed a method for attacking hidden tasks in a multitask classifier, considering a threat model where the knowledge available to the attacker varies for each task. The results of the experiments demonstrated that the proposed method is more effective for hidden tasks compared to existing methods while mitigating the impact on other tasks. As the use of pre-trained models in model building is becoming commonplace, it is necessary to be aware of CF attacks and CF delta attacks as one of the risks of dealing with publicly available pre-training models as they are.

For future work, analyzing the reasons why CF delta attack remains effective even when the headers are nonlinear layers, such as examining changes in feature vectors, such as did in \cite{ilyas2019adversarial}, could provide insights for further improving attacks. Moreover, exploring how to defend against such an attack would also be a valuable topic. Whether defenses against non-hidden tasks \cite{shafahi2019adversarial, he2017adversarial} can work against attacks designed for hidden target task, as discussed in this study, remains a question that can be explored.


\newpage
\bibliographystyle{ACM-Reference-Format}
\bibliography{z_bib}


\end{document}

%% file: defs.tex
\usepackage{xparse}
\newcommand{\bnm}{\begin{newmath}}
\newcommand{\enm}{\end{newmath}}

\newcommand{\bea}{\begin{eqnarray*}}%
\newcommand{\eea}{\end{eqnarray*}}%

\newcommand{\bne}{\begin{newequation}}
\newcommand{\ene}{\end{newequation}}

\newcommand{\bal}{\begin{newalign}}
\newcommand{\eal}{\end{newalign}}

\newenvironment{newalign}{\begin{align}%
\setlength{\abovedisplayskip}{4pt}%
\setlength{\belowdisplayskip}{4pt}%
\setlength{\abovedisplayshortskip}{6pt}%
\setlength{\belowdisplayshortskip}{6pt} }{\end{align}}

\newenvironment{newmath}{\begin{displaymath}%
\setlength{\abovedisplayskip}{4pt}%
\setlength{\belowdisplayskip}{4pt}%
\setlength{\abovedisplayshortskip}{6pt}%
\setlength{\belowdisplayshortskip}{6pt} }{\end{displaymath}}

\newenvironment{newequation}{\begin{equation}%
\setlength{\abovedisplayskip}{4pt}%
\setlength{\belowdisplayskip}{4pt}%
\setlength{\abovedisplayshortskip}{6pt}%
\setlength{\belowdisplayshortskip}{6pt} }{\end{equation}}

\newcounter{ctr}

\newcounter{mytable}
\def\mytable{\begin{centering}\refstepcounter{mytable}}
\def\endmytable{\end{centering}}

\newcounter{myfig}
\def\myfig{\begin{centering}\refstepcounter{myfig}}
\def\endmyfig{\end{centering}}

\newlength{\saveparindent}
\newlength{\saveparskip}
\newcommand{\E}{{\rm I\kern-.3em E}}

\newcommand{\secref}[1]{\mbox{Section~\ref{#1}}}

\renewcommand{\eqref}[1]{\mbox{Equation~(\ref{#1})}}

\def \part {part}

\renewcommand{\paragraph}[1]{\vspace*{6pt}\noindent\textbf{#1}\;}

\def \blackslug{\hbox{\hskip 1pt \vrule width 4pt height 8pt
    depth 1.5pt \hskip 1pt}}
\def \qed{\quad\blackslug\lower 8.5pt\null\par}

\newcounter{mynote}[section]

\newcommand\ignore[1]{}

\newcounter{rcnote}[section]

\newcounter{mrnote}[section]

\newcounter{fknote}[section]

\newcounter{anote}[section]

\DeclareMathSymbol{\mlq}{\mathord}{operators}{``}
\DeclareMathSymbol{\mrq}{\mathord}{operators}{`'}

\newcommand{\rhf}[2]{R_{f, \gamma}}

\DeclareDocumentCommand{\edist}{o o}{
  \ensuremath{
    \IfNoValueTF{#1}{{d}}{{\sf d}(#1,#2)}
  }
}

\newcommand{\olrk}[1]{\ifx\nursymbol#1\else\!\!\mskip4.5mu plus 0.5mu\left(\mskip0.5mu plus0.5mu #1\mskip1.5mu plus0.5mu \right)\fi}

\NewDocumentCommand{\indseq}{ O{1} O{r} }{{#1}\ldots {#2}}
